\documentclass[natbib,referee]{svjour3}

\usepackage[english]{babel}
\usepackage{amssymb,amsmath}
\usepackage{graphicx,subfigure}
\usepackage{mathptmx}
\usepackage{stmaryrd}
\usepackage{url}
\usepackage{float}
\usepackage{color}

\spnewtheorem{algorithm}{Algorithm}{\bf}{\rm}

\newcommand{\Sec}[1]{Sec.~\ref{#1}}
\newcommand{\Eq}[1]{Eq.~(\ref{#1})}

\newcommand{\Tab}[1]{Tab.~\ref{#1}}

\newcommand{\TT}[1]{\text{\tt #1}}
\newcommand{\id}{\mathrm{id}}

\newcommand{\last}{\mathrm{last}}
\newcommand{\cg}{\mathrm{cg}}
\newcommand{\mng}[1]{\llbracket #1 \rrbracket}

\journalname{Journal of Logic, Language, and Information}

\bibpunct{(}{)}{,}{}{}{;}

\begin{document}


\title{Machine Semiotics}
\titlerunning{Machine Semiotics}

\author{
    Peter beim Graben \and
    Markus Huber-Liebl \and \\
    Peter Klimczak \and
    G\"unther Wirsching
}

\authorrunning{beim Graben, Huber-Liebl, Klimczak, \& Wirsching}

\institute{
    Peter beim Graben,
    Bernstein Center for Computational Neuroscience Berlin, Germany
    \email{peter.beimgraben@b-tu.de}
    \and
    Markus Huber-Liebl,
    Kommunikationstechnik, Brandenburgische Technische
    Universität Cottbus-Senftenberg, Cottbus, Germany
    \and
    Peter Klimczak,
    Angewandte Medienwissenschaften, Brandenburgische Technische
    Universität Cottbus-Senftenberg, Cottbus, Germany
    \and
    G\"unther Wirsching,
    Mathematik und Statistik, Katholische Universit\"at
    Eichst\"att-Ingolstadt, Eichst\"att, Germany
}

\date{\today}
\maketitle

\begin{abstract}
Recognizing a basic difference between the semiotics of humans and machines presents a possibility to overcome the shortcomings of current speech assistive devices. For the machine, the meaning of a (human) utterance is defined by its own scope of actions. Machines, thus, do not need to understand the conventional meaning of an utterance. Rather, they draw conversational implicatures in the sense of (neo-)Gricean pragmatics. For speech assistive devices, the learning of machine-specific meanings of human utterances, i.e. the fossilization of conversational implicatures into conventionalized ones by trial and error through lexicalization appears to be sufficient. Using the quite trivial example of a cognitive heating device, we show that --- based on dynamic semantics --- this process can be formalized as the reinforcement learning of utterance-meaning pairs (UMP).
\end{abstract}


\section{Introduction}
\label{sec:intro}

The conventional meaning of an utterance, such as ``I am going to grandma'' results compositionally from the meaning of its linguistic constituents, phrases and eventually single words. For the given example, we consider a speaker to whom the personal pronoun ``I'' is referring to. Moreover, this speaker announces her departure from the current place using present progressive tense (``am going'') with a locative goal (``to grandma''), such that the conventional meaning of the utterance can be formally circumscribed somehow as EVENT: go(AGENT: speaker, TIME: now, GOAL: grandmother(RELATION: speaker)). However, such semantic descriptions are largely underspecified in general \citep{Blutner98b}. Therefore, the pragmatic sense of an utterance also depends crucially upon its respective context, focus, and presuppositions. In our example, focusing upon departure from the current place may at least indicate the hearer's privacy.

Considering the following contextualization: ``{}`I am going to grandma', said Little Red Riding Hood'', the speaker might be a mother telling her child Grimm's fairy tale. Being an expert of Grimm's fairy tales, the child may immediately ask her mother about the later fate of Bad Wolf. In another different context, the speaker could be actually that child, expecting a substantial money gift when visiting her grandmother.

In the examples above, the hearer of an utterance has to draw inferences from what the speaker has said. In some cases, the hearer's inferences are logically implied by the conventional meaning of an utterance. In order to distinguish logical implications from other types of pragmatic inferences, \citet{Grice89} coined the term \emph{implicature}\footnote{
    Originally published as \emph{Logic and Conversation} (1975).
}
stating ``[\dots] the implicature is not carried by what is said, but only by the saying of what is said, or by `putting it that way''' \cite[p.~39]{Grice89}.

\Citet{Grice89} elaborated his theory of pragmatic implicature by presenting some normative principles and maxims of cooperative conversation that are summarized in \Tab{tab:grice}.

\begin{table}[H]
  \centering
   \begin{description}
    \item[\bf Cooperative Principle] Make your conversational contribution such as is required, at the stage at which it occurs, by the accepted purpose or direction of the talk exchange in which you are engaged.
    \item[{\bf Maxims of Quantity} (Informativity)] ~
        \begin{enumerate}
          \item Make your contribution as informative as is required.
          \item Do not make your contribution more informative than is required.
        \end{enumerate}
    \item[{\bf Maxim of Quality} (Veridicality)] Try to make your contribution one that is true.
        \begin{enumerate}
          \item Do not say what you believe to be false.
          \item Do not say that for which you lack adequate evidence.
        \end{enumerate}
    \item[{\bf Maxim of Relation} (Relevance)] Be relevant.
    \item[{\bf Maxim of Manner} (Perspicuity)] Be perspicuous.
        \begin{enumerate}
          \item Avoid obscurity of expression.
          \item Avoid ambiguity.
          \item Be brief.
          \item Be orderly.
        \end{enumerate}
\end{description}
  \caption{The Cooperative Principle and Conversation Maxims after \citet[p.~26f]{Grice89}.}\label{tab:grice}
\end{table}

In \Tab{tab:grice} we present the Cooperative Principle and the Conversation Maxims, together with a characteristic keyword, not given by \citet{Grice89}. The maxims of Quantity are concerned with the informativity of an utterance. The first Quantity Maxim demands that an utterance must be sufficiently informative to be understood by the hearer. By contrast, the second Quantity Maxim reduces the speaker's information encoding effort to a necessary minimum. The Maxims of Quality can be regarded as maxims of veridicality, preventing lies or the dissemination of \emph{fake news}. Correspondingly, the Maxims of Relation and of Manner require the speaker to utter only issues of relevance in a perspicuous way. Obviously, these Conversation Maxims are not independent from each other. If the speaker's utterance is to much verbose, it could also be more informative as required, containing irrelevant or obscure messages as well.

According to \citet{Grice89}, a speaker has to comply with these normative principles for successful communication. However, she might also intentionally deviate in one or another respect from one or the other maxim as long as the overall Cooperative Principle is maintained. In this case, the hearer is encouraged by the speaker to draw an implicature from what has been communicated.

Specifically, \citet{Grice89} distinguished between two kinds of pragmatic implicatures. On the one hand, he introduced \emph{conventional implicatures} that could be derived from the conventional meaning of the utterance. On the other hand, he defined \emph{conversational implicatures} that are non-compositional inferences from what the speaker could have possibly meant by departing from the Conversation Maxims. Although the former concept of conventional implicature has been quite controversial in the linguistic literature (cf.~\citet{Bach99}), it found a satisfactory revision in recent time by \citet{Potts03} and, among others, \citet{VenhuizenBosEA14}. Applied to the examples above, one could state that the Grimm expert child draws a conventional implicature about Bad Wolf, as Little Red Riding Hood provides the semantic anchor for pragmatic enrichment in the sense of \citet{VenhuizenBosEA14}. By contrast, the child visiting her grandmother to expect a money gift, invites her mother to draw the conversational implicature that she is actually visiting her grandmother to obtain a little pocket money through her utterance.

Yet from now on let us assume that the addressee of the speaker's utterance is an Artificial Intelligence \citep{Wheatman14}. A particular instance of such a machine has been discussed by \citet{KlimczakWolffLindemannEA14} and \citet{HuberEA18} as their \emph{cognitive heating}, namely a speech-assistive heating device in a smart home environment. The cognitive heating is a cognitive dynamical system \citep{Haykin12}, i.e., a \emph{cognitive agent}. This agent is embedded into its environment, interacting through a perception-action cycle (PAC) \citep{Uexkull82, Young10}, where particular sensors provide information about the state of the environment that is relevant for the agent's behavior. In terms of the ecological theory of meaning of \citet{Uexkull82}, this perceptual arc couples the agent to its subjective \emph{merkwelt}. On the other hand, the agent is able to act on its subjective \emph{wirkwelt} by means of specialized effectors along its actuator arc. Sensory merkwelt and operative wirkwelt together form the agent's \emph{umwelt} as its subjective universe.

In the case of the cognitive heating, perception comprises sensation of temperature. But more importantly, we crucially assume elaborated speech recognition capacities, enabling the agent to successfully process and classify speech signals in order to compute a symbolic representation of user utterances \citep{GravesMohamedHinton13, SundermeyerNeySchluter15}. Thus the merkwelt of the cognitive heating comprises its fundamental speech recognition capability, yet without any semantic understanding, logical reasoning, or pragmatic world knowledge. Moreover, in its most simple case considered here, action is restricted to either switching on or off the heating furnace, thereby comprising the agent's wirkwelt.

Given this kind of perception-action cycle, we address the problem of \emph{semiosis}, or \emph{symbol grounding} \citep{Harnad90, Posner93}: How does a cognitive agent assign meaning to symbolic data during language acquisition? For the cognitive heating this simply means to assign an operation mode (to heat or not to heat) to any possible utterance, such as ``I am going to grandma''. Our particular solution, called \emph{Machine Semiotics}, is inspired by fundamental insights from constructivism \citep{MaturanaVarela98, Forster03}. According to \citet{MaturanaVarela98}, linguistic behavior of a sender aims at changing (linguistic) behavior of the receiver (also cf.  \citet{Grice89, Dennett89g, Posner93}). Applied to the cognitive heating, we propose a simple reinforcement algorithm \citep{Skinner57, SuttonBarto18} that changes the agent's behavior upon any non-empty utterance, where the meanings of utterances become learned during semiosis. By contrast, if the machine's operator remains silent, our algorithm interprets this kind of linguistic behavior as consent with the current operation mode.

Employing Grice' \citeyearpar{Grice89} theory of pragmatic implicature to the devised intelligent machine leads to interesting consequences. First of all, we do not postulate any kind of compositional semantics the machine is required to process. Therefore, the agent has no access to the user's conventional meanings. Hence, it is also not capable to draw conventional implicatures unless conventional meaning had been acquired through the envisaged semiotic learning process. Second, the meaning domain of the machine results solely from its range of possible actions. In case of a cognitive heating, there are only two possible actions: either turn on or turn off the heater (deliberately simplifying from temperature tuning here). In other words, the possible actions are either maintaining or changing the current working state of the device. Thus, the machine interprets the utterances of a user always as imperatives with respect to its own faculties \cite[p.~123]{Grice89}.

How would the cognitive heating understand the user's utterance ``I am going to grandma''? It depends on the current working state. Assume that the user makes that utterance in case the heater is currently turned on, with the desire that no heating power is required during her absence. Because the machine has no access to the conventional meaning of what the user says, it considers any utterance relevant by relating it to its own action space; this provides our third essential characteristic of Machine Semiotics. Now, because the heater is currently turned on, the machine turns it off as a reaction to the utterance. Yet this is (under normal circumstances) exactly what the user means:  ``Since I am going to grandma, I don't need heating power this afternoon''. Hence, the machine correctly draws the conversational implicature, intended by the user. Fourth, the reinforcement learning algorithm enables the machine to memorize the successfully inferred conversational implicatures that become \emph{conventionalized} \cite[p.~39]{Grice89} through \emph{lexicalization}. In the related framework of optimality-theoretic pragmatics, this process has been called \emph{fossilization} \citep{BlutnerZeevat09}. Thus, our machine learning approach could be related to linguistic research on diachronic language evolution.

For our method deploys recent developments of formal pragmatics and computer science, we present a brief overview about neo-Gricean optimization in the next paragraphs.

As already mentioned above, Grice' \citeyearpar{Grice89} Conversation Maxims given in \Tab{tab:grice} are not independent from each other. Therefore, scholars of the neo-Gricean movement, such as \citet{AtlasLevinson81} and \citet{Horn84}, achieved a substantial simplification of the Conversation Maxims. Inspired by Zipf's \citeyearpar{Zipf49} \emph{Principle of Least Effort}, \citet{Horn84} reduced the four categories of Conversation Maxims to only a pair of two, referred to as the Principle of Quantity (Q Principle) and the Principle of Relation (R Principle).\footnote{
    Called Principle of Informativity (I Principle) in a related work by \citet{AtlasLevinson81}.
}
Table \ref{tab:horn} summarizes Horn's Principles \citeyearpar{Horn84} below.

\begin{table}[H]
  \centering
   \begin{description}
    \item[\bf Q Principle (hearer-based)] ~
    \begin{enumerate}
          \item Make your contribution sufficient.
          \item Say as much as you can (given R).
        \end{enumerate}
    \item[\bf R Principle (speaker-based)] ~
    \begin{enumerate}
          \item Make your contribution necessary.
          \item Say no more than you must (given Q).
        \end{enumerate}
\end{description}
  \caption{The Q and R Principle \citep[p.~13]{Horn84}.}\label{tab:horn}
\end{table}

Under the assumption that both partners in a conversation behave cooperatively (obeying Grice' Principle of Cooperation), they are necessarily committed to the Maxims of Quality \citep[p.~12]{Horn84}. Then, the Q Principle in \Tab{tab:horn} comprises the first Maxim of Quantity \citep[p.~14]{Horn84} and the first two Maxims of Manner \citep[p.~6]{BlutnerZeevat09}. Likewise, the R Principle captures the second Maxim of Quantity, the Maxim of Relation \citep[p.~14]{Horn84} and also the last two Maxims of Manner \citep[p.~6]{BlutnerZeevat09}. \Citet[p.~12]{Horn84}, emphasized that those principles express different aspects of optimization in his study. Considering the R Principle first, it could be related to Zipf's Principle of Least Effort \citep{Zipf49}, in that a speaker tries to minimize the complexity of her utterance, e.g. expressed by the time demand of speech production. By contrast, the Q principle obliges the speaker to maximize the informativity of her utterance such that the hearer is able to minimize her own processing effort. Interestingly, both principles are recursively invoking each other, as the Q Principle refers to the R Principle, and vice versa.

Based on Horn's Conversation Principles \citeyearpar{Horn84}, \citet{Blutner98b} was able to provide a first mathematical formalization of the Gricean maxims in terms of complexity and informativity of the (conventional) meaning of an utterance. Later, his definitions gave rise to the foundation of optimality-theoretic pragmatics \citep{Jager02, Blutner06, BlutnerZeevat09, Benz09, HawkinsFrankeEA22}. Since we rephrase Blutner's \citeyearpar{Blutner98b} formal codification in terms of Machine Semiotics subsequently, we only present an informal account in this section.

Under the assumption that complexity and informativity measures of an utterance are well-defined, we consider two complementary decision situations. On the one hand, the speaker, having some meaning in mind, wants to express this mental representation in the most economic way towards the hearer. If the same meaning can be expressed by two different utterances, one less complex than the other, the R Principle suggests the speaker to chose the expression with lower complexity. On the other hand, the hearer has to decode the given utterance into the indented meaning. This is highly facilitated if the expression has as little divergent meanings as possible. Thus, the speaker is exhorted by the Q Principle to decide for an utterance that is as little ambiguous as possible, or, in other words, to maximize the hearer's informativity \citep{Blutner98b}. Moreover, \citet{Blutner98b} provided also a formally rigorous account to Grice' Maxims of Quality in \Tab{tab:grice} by means of a propositional common ground shared by speaker and hearer. This idea is also further exploited subsequently.

The article is structured as follows. In \Sec{sec:meth} we present the methodological techniques used in our study, which are dynamic semantics in \Sec{sec:dynsem}, formal pragmatics in \Sec{sec:fprag}, and reinforcement learning in \Sec{sec:rforlear}. We suggest a learning algorithm for Machine Semiotics and prove some of its properties. In particular, we demonstrate soundness of the algorithm and its convergence when the user complies with the Conversation Maxims. In \Sec{sec:results} we apply the algorithm to two example scenarios, one for the simple heating device with binary decision space, and another one for a three-valued decision space, where we show how a mental lexicon of machine-relevant utterance-meaning pairs can be acquired. In the final \Sec{sec:disc} we discuss our approach in a broader semiotic context.

\section{Methods}
\label{sec:meth}

We present a promising approach for solving the technical problem of fossilization in a machine learning context. The suggested solution rests upon the fact that the only relevant meanings are given for a machine by a range of its possible actions to which we refer to as the \emph{action space} in the following. Our methods combine dynamic semantics \citep{Gardenfors88, GroenendijkStokhof91, Kracht02, Graben14a} and formal pragmatics \citep{Grice89, AtlasLevinson81, Horn84, Blutner98b, Blutner06, HawkinsFrankeEA22} with reinforcement learning \citep{Skinner57, SuttonBarto18}.

\subsection{Dynamic Semantics}
\label{sec:dynsem}

We describe a general cognitive agent (the machine, in the present setting) as a \emph{cognitive dynamical system} $\Sigma_s = (X_s, \Phi_s^t)$, where the set of actions $X_s$ comprises the \emph{state space} of the particular cognitive agent under consideration \citep{Haykin12}, indicated by the index $s$ (to be omitted henceforth), while the \emph{flow} $\Phi_s^t : X_s \to X_s$ is given through a one-parameter semigroup homomorphism with the property $\Phi_s^t \circ \Phi_s^r = \Phi_s^{t + r}$, with an additive semigroup $(G, +)$ \citep{AtmanspacherGraben07, Robinson99}. The parameter $t \in G$ is conventionally interpreted either as discrete ($G = \mathbb{N}_0$) or continuous ($G = \mathbb{R}_0^+$) time, respectively. Another viable interpretation of $t$ as an integer or real random number, leads to stochastic dynamics. Note that a proper dynamic group $(G, +)$ describes a deterministic dynamics that would not be compatible with such a stochastic description. Preparing an \emph{initial condition} $x_0 \in X_s$ and letting the dynamics evolve over some time span $0 \le t \le \tau$, yields a \emph{trajectory} $T = \{ x(t) \in X_s | x(t) = \Phi_s^t(x_0) \, , \,  0 \le t \le \tau \}$ in state space.

For the sake of simplicity, we assume that the state space $X_s$ is either a finite set or at most recursively enumerable. Moreover, we demand the system to be \emph{ergodic}, which means here that for any two states, $x, y \in X_s $ there is a parameter $t \in G$, such that $y = \Phi_s^t(x)$. For an ergodic dynamical system, any state is connected with any other state through a trajectory of sufficient length in state space. One particular example would be a Markov chain with irreducible transition matrix \citep{AtmanspacherGraben07}.

Further, we suggest reinforcement learning as discussed by \citet{Skinner57} for the machine's acquisition of verbal semantics. Skinner argued that the meaning of a linguistic utterance, called \emph{verbal behavior} $b$, is not directly related to either the utterance nor to its linguistic constituents. Rather he stated that verbal behavior $b$ acts as a function on \emph{pragmatic antecedents} $a \in X_s$, mapping them into \emph{pragmatic consequents} $c \in X_s$; what has been called `ABC' schema. According to Skinner, the meaning of $b$, conventionally denoted $\mng{b}$, is a partial function from antecedents to consequents, or formally
\begin{equation}\label{eq:abc}
  \mng{b}: a \mapsto c
\end{equation}
with $a, c \in X_s$ as possible actions.

In this article, we argue that Skinner's ABC schema, as explained above, is compatible with the ideas of dynamic semantics \citep{Gardenfors88, GroenendijkStokhof91, Kracht02, Graben14a} and of formal pragmatics \citep{Grice89, AtlasLevinson81, Horn84, Blutner98b, Blutner06, HawkinsFrankeEA22}, making it a natural candidate for reinforcement learning of Machine Semiotics, in contrast to some claims of \citet{Chomsky59}.

In experimental psychology, measurement data are associated to antecedents $a$, behavior $b$ and consequents $c$ in such a way that an experimental situation is described by a joint probability distribution $p(a, b, c)$. Then, the meaning of (verbal) behavior can be described in terms of Bayesian conditionalization $p(a, b, c) = p(c|a, b) p(b|a) p(a)$ where $p(a)$ is the apriori probability distribution of antecedents $a \in X_s$, $p(b|a)$ is the conditional probability that the agent encounters event $b$ (the behavior of another agent --- e.g. its utterance) in situation $a$, and $p(c|a, b)$ expresses the conditional probability of the consequent $c \in X_s$, that is the agent's decision in the situation complex $(a, b)$. In dynamic semantics, this idea is known as Bayesian updating of epistemic states (cf. \citet{Williams80, Gardenfors88, BenthemGerbrandyKooi09}, and notably \citet{Benz09, Benz16}, as well as \citet{HawkinsFrankeEA22} as elaborated accounts of Bayesian pragmatics). Here, an \emph{epistemic state} or likewise a \emph{belief state} or a \emph{semantic anchor} \citep{WirschingLorenz13}, is a subjective reference to a subjective reality, i.e. an \emph{umwelt} in the sense of \citet{Uexkull82}. In the context of Bayesian descriptions, it becomes a (subjective) probability distribution over the subjective action space $X_s$ of the cognitive agent $s$ \citep{Jaynes57a, Jaynes57b}. This also connects our approach to reinforcement learning as a machine learning technology \citep{SuttonBarto18}. However, in our current exposition we refrain from probabilistic intricacies in the first place by formally modeling the acquisition of Machine Semiotics through utterance meaning pairs (UMP) \citep{KwiatkowskiGoldwater12, WirschingLorenz13, GrabenEA19b}.

Following \citet{Skinner57}, we assign to each compound of the ABC schema, one epistemic state that could change from time to time by dynamic updating. As mentioned above, an epistemic state is a subjective reference to a subjective reality, i.e. an \emph{umwelt}. For our first most simple example, the subject is a heating device as a cognitive agent. Its subjective reality, i.e. is action space $X_s$, consists of only two operation modes: to heat or not to heat. As epistemic states, we therefore consider two subjective references, or beliefs: (1) ``I operate in heating mode'' ($H$), and (2) ``I operate in non-heating mode'' ($\neg H$). Thus, we have
\begin{equation}\label{eq:actions}
  X = \{ \neg H, H \}
\end{equation}
and also relations
\begin{equation}\label{eq:rel}
  F \subset X \times X = \{ (\neg H, \neg H), (\neg H, H),  (H, \neg H), (H, H) \}
\end{equation}
that become utilized subsequently.

Next, we consider \emph{utterance-meaning pairs} (UMP) \citep{KwiatkowskiGoldwater12, WirschingLorenz13, GrabenEA19b} (also called \emph{form-interpretation pairs} in the context of formal pragmatics \citep{Jager02, Blutner06}).

\begin{definition}\label{def:ump}
An UMP $U$ is an ordered pair of a phonetically or graphemically encoded token of speech $u$ (i.e. verbal behavior $u = b$) and its \emph{meaning} $\mng{u}$,
\begin{equation}\label{eq:ump}
  U = (u, \mng{u}) \:.
\end{equation}
\end{definition}

In dynamic semantics, meaning is defined as a partial function on epistemic states \citep{Gardenfors88, GroenendijkStokhof91, Kracht02, Graben14a}. And according to \citet{Skinner57} epistemic states are antecedents $a$ and consequents $c$ that result from applying linguistic behavior $u = b$ onto $a$, i.e.
\begin{equation}\label{eq:abc2}
  c = \mng{u}(a) \:.
\end{equation}
In the sequel, we essentially follow \citet{GroenendijkStokhof91} and \citet{Kracht02} and define this partial function by explicitly enumerating the corresponding relations $F \subset X \times X$.
\begin{definition}\label{def:mng}
The meaning $\mng{u}$ of an utterance $u$ is a set of ordered antecedent-consequent pairs
\begin{equation}\label{eq:mean}
    \mng{u}  = F = \{(a_1, c_1), (a_2, c_2), \dots (a_m, c_m) \} \:,
\end{equation}
where $(a_i, c_i) \in F$ such that $a_i = a_j \Rightarrow c_i = c_j$ for $i \ne j$ in order to make $\mng{u}$ a well-defined partial function on $X$.
\end{definition}

With respect to our training algorithm below, we have to take the intrinsic dynamics of the system into account. Therefore, we define the dynamic updating of any epistemic state $x \in X$ as follows.

\begin{definition}\label{def:mngup}
The meaning function $\mng{u}$ acts onto an arbitrary epistemic state $x \in X$ through
\begin{equation}\label{eq:meanup}
    y = \mng{u}(x) = \begin{cases}
        c & \text{if } (a, c) \in \mng{u} \text{and } x = a \\
        \Phi^\tau(x) & \text{else} \:.
    \end{cases}
\end{equation}
\end{definition}
In definition \ref{def:mngup} the parameter $\tau \in G$ is suitably chosen in such a way that different utterances map one antecedent state onto different consequent states. This could be achieved, e.g., by letting $\tau$ be the duration of the utterance $u$, or, perhaps more appropriately, as a random number for a stochastic Markov chain.

Additionally, have to endow the meaning relation with a total order.
\begin{definition}\label{def:to}
The meaning relation $F = \mng{u}$ is equipped with a total order $(a_p, c_p) \preceq (a_q, c_q)$ if $p \le q$ for indices $p, q \in \mathbb{N}_0$. Hence, $F$ possesses a maximum
\[
    \max F = (a_m, c_m)
\]
with $m$ the largest index in \Eq{eq:mean}.
\end{definition}

One important notion of dynamic semantics is that of \emph{acceptance} \citep{Gardenfors88, Graben14a}.
\begin{definition}\label{def:acc}
An epistemic state $x \in X$ is said to accept the meaning $\mng{u}$ of an utterance $u$, if $x$ is a fixed point of $\mng{u}$, i.e.
\begin{equation}\label{eq:acc}
    x = \mng{u}(x) \:,
\end{equation}
conversely, $\mng{u}$ is said being accepted by state $x \in X$.
\end{definition}

\subsection{Formal pragmatics}
\label{sec:fprag}

In order to present our ideas of successful ``conversation'' between a user and the cognitive machine, we adopt some concepts from formal pragmatics with are appropriately adapted to the aims of Machine Semiotics \citep{Grice89, AtlasLevinson81, Horn84, Blutner98b, Blutner06, HawkinsFrankeEA22}. Here we follow essentially the original ideas of \citet{Blutner98b} before they have been successfully applied to the framework of optimality-theoretic pragmatics \citep{Blutner98b, Jager02, Blutner06}.

First, our definition \ref{def:mngup} leads to a natural distinction between conventional \citep{Grice89, Potts03, VenhuizenBosEA14} and conversational implicatures \citep{Grice89, Blutner98b, Blutner06, Benz09} in dynamic semantics.

\begin{definition}\label{def:implic}
Let $\mng{u}$ be a meaning function, piecewise defined as in \Eq{eq:meanup},
\[
    y = \mng{u}(x) = \begin{cases}
        c & \text{if } (a, c) \in \mng{u} \text{and } x = a \\
        \Phi^\tau(x) & \text{else} \:.
    \end{cases}
\]
On the one hand, we refer to the first branch as to a \emph{conventional implicature} for $(a, c) \in \mng{u}$ is lexically anchored \citep{VenhuizenBosEA14}. On the other hand, the second branch is interpreted as a \emph{conversational implicature}, because the machine has to draw an inference according to its intrinsic dynamics without explicitly accessing any conventional meaning.
\end{definition}

Inspired by \citet{Blutner98b}, we next suggest what counts as speaker- and as hearer-optimality in the context of Machine Semiotics.

\begin{definition}\label{def:cod}
Let $(u, \mng{u})$ be an UMP. The \emph{complexity} of $(u, \mng{u})$ is given by the length
\begin{equation}\label{eq:cmplx}
  k = |u| \in \mathbb{R}_0^+
\end{equation}
of the given phonetic or graphemic encoding of the utterance $u$. We call an UMP $(u_1, \mng{u_1})$ \emph{less complex} than a second UMP $(u_2, \mng{u_2})$, if $|u_1| < |u_2|$.
\end{definition}
Note that in the case of phonetic encoding, the length of the utterance could be identified with its real-time duration $\tau$.

Moreover, we follow \citet{AtlasLevinson81} with their understanding of informativity.

\begin{definition}\label{def:inf}
Let $(u_1, \mng{u_1}), (u_2, \mng{u_2})$ be two UMP. The UMP $(u_1, \mng{u_1})$ is said to be \emph{more informative} than the UMP $(u_2, \mng{u_2})$ if
\begin{equation}\label{eq:info}
  \mng{u_2} \subset \mng{u_1}
\end{equation}
for the corresponding meaning relations.
\end{definition}

Now, we are able to reformulate Blutner's \citeyearpar{Blutner98b} Q and R (I) principles of bidirectional optimization.

\begin{definition}[Q Principle]\label{def:bq}
A UMP $(u, \mng{u})$ is said to be \emph{hearer optimal}, if there is no other UMP $(u', \mng{u'})$ that is simultaneously speaker optimal and more informative than $(u, \mng{u})$.
\end{definition}

\begin{definition}[R Principle]\label{def:br}
A UMP $(u, \mng{u})$ is said to be \emph{speaker optimal}, if there is no other UMP $(u', \mng{u'})$ that is simultaneously hearer optimal and less complex than $(u, \mng{u})$.
\end{definition}

Both principles are formulated for speech production here in contrast to Blutner's \citeyearpar{Blutner98b} original definition focussing on speech comprehension. However, as \citet[p.~132]{Blutner98b} emphasized, both principles are formally symmetric such that a change of perspective results from interchanging the symbols R (I) and Q, respectively. As discussed in the introduction, these principles describe decision problems for the speaker. In the case of definition \ref{def:bq}, when the speaker wants to convey a conceived meaning to the hearer, she has to decide between two UMPs $(u_1, \mng{u_1}), (u_2, \mng{u_2})$ where no one is more informative than the other. Then Principle R suggests to minimize complexity. In the other case of definition \ref{def:br}, when no UMP is less complex than the other one, principle Q advises the speaker to maximize the hearers informativity. Note that both principles are rigorous formalizations of the respective principles given by \citet{Horn84} [\Tab{tab:horn}]. Further note, that the same kind of circularity appears in definitions \ref{def:bq} and \ref{def:br} as in  Horn's \citeyearpar{Horn84} formulation. This apparent circularity becomes resolved in a properly recursive definition inspired by \citet{Jager02} and \citet{Blutner06}.

\begin{definition}[Super Optimality]\label{def:bbo}
A UMP $(u, \mng{u})$ is said to be \emph{super-optimal}, if
    \begin{enumerate}
        \item {\bf Q Principle} There is no other UMP $(u', \mng{u'})$ that is simultaneously super-optimal and more informative than $(u, \mng{u})$.
        \item {\bf R Principle} There is no other UMP $(u', \mng{u'})$ that is simultaneously super-optimal and less complex than $(u, \mng{u})$.
    \end{enumerate}
\end{definition}

\citet{Blutner98b} also formalized Grice' Maxims of Quality (Veridicalilty) in terms of a common ground of shared propositions. To this aim, we distinguish between the meaning relations of two agents, the speaker, $s$, and the hearer, $h$, in such a way that $\mng{u}_s$ denotes the meaning of utterance $u$ for the speaker, and $\mng{u}_h$ comprises the meaning of $u$ for the hearer.

\begin{definition}[Common Ground]\label{def:cogro}
Let $u$ be an utterance such that $\mng{u}_s$ is the meaning relation of a speaker and $\mng{u}_h$ that of a hearer. We define the common ground of $u$ as the intersection
\begin{equation}\label{eq:cg}
    \cg(u)_{s,h} = \mng{u}_s \cap \mng{u}_h  
\end{equation}
which is well-defined provided that none of the meaning relations is empty, otherwise, the common ground of speaker and hearer becomes the empty set.
\end{definition}

Loosely following \citet{Blutner98b} again, we make the following ansatz.\footnote{
    \citet{Blutner98b} combined his codification of the Quality Maxims with those of Horn's \citeyearpar{Horn84} Quantity and Relation Principles. We refrain from this complication in the Machine Semiotic and shall postpone it to later investigation.
}

\begin{definition}[Quality Principle 1]\label{def:q1}
If the speaker wants the hearer to apply a state transition $a \mapsto c$ by uttering $u$, i.e. $(a, c) \in \cg(u)_{s,h}$, she must not utter $u'$ when $(a, c) \not\in \cg(u')_{s,h} \ne \emptyset $ though.
\end{definition}
A violation of this principle reflects the case that the speaker makes an utterance that she believes to be false, where ``falsehood'' is dynamically understood as an undesired state transition. Thus, we assume, that the speaker desires a transition $a \mapsto c$, but instead of uttering $u$ appropriately, knowing that $u$ entails the desired transition, for $u$ belongs to her common ground with the hearer, she voluntarily decides to utter $u'$, also knowing that $u'$ does not lead to the desired transition. Hence, she says what she believes to be false.

Similarly, we present another definition.
\begin{definition}[Quality Principle 2]\label{def:q2}
If the speaker wants the hearer to apply a state transition $a \mapsto c$ by uttering $u$, i.e. $(a, c) \in \cg(u)_{s,h}$, she must not utter $u'$ when $(a, c) \not\in \cg(u')_{s,h} = \emptyset $ though.
\end{definition}
Here, we have a slightly different situation. In the case of violating Quality Principle 1, the speaker utters $u'$, knowing that it leads to a ``false'' state description. In the second case, she knows that the transition $a \mapsto c$ is not defined for the hearer, thus uttering $u'$ means that she lacks sufficient evidence.

\subsection{Reinforcement Learning}
\label{sec:rforlear}

The aim of our fossilization algorithm is the construction of a \emph{mental lexicon} of UMPs
\begin{equation}\label{eq:menlex}
  M_T = \{U_1, U_2, \dots, U_T \}
\end{equation}
after training time $T$ through reinforcement learning \citep{Skinner57, SuttonBarto18, GrabenEA19b}, where
\[
	U_k = (u_k, \mng{u_k})
\]
is an UMP encountered at iteration $k$ with meaning
\begin{equation}\label{eq:mean2}
    \mng{u_k}  = \{(a_{k1}, c_{k1}), (a_{k2}, c_{k2}), \dots (a_{k{m_k}}, c_{k{m_k}}) \} \:,
\end{equation}
with $(a_{ki}, c_{ki}) \in F$. Iteration time is regarded as being discretized. To this end, we associate to each instance of time $k$ a corresponding ABC schema
\begin{equation}\label{eq:abciter}
  a_k \stackrel{b_k}{\longmapsto} c_k
\end{equation}
in order to model the verbal behavior of the user and the resulting machine's actions.

Starting from a \emph{tabula rasa} state at iteration $k = 0$, we initialize our system with empty lexicon $M_0 = \emptyset$ and antecedent state $a_0 \in X$. To the behavior $b_0$ we associate the given utterance $u_0$. Additionally, we assume that the cognitive heating displays its current working (epistemic) state either visually (continuously) and/or acoustically to the user after every state transition, making the machine states \emph{observable} to the user \citep{RussellNorvig10}. Every further iteration is initialized with the consequent of the previous iteration as antecedent state,
\begin{equation}\label{eq:iniup}
    a_{k+1} \gets c_k
\end{equation}
for $k \in \mathbb{N}_0$.

For the sake of convenience, we introduce the following notations. Let $U = (u, \mng{u})$ be a UMP. We define two \emph{projectors} $\mathrm{P}, \mathrm{Q}$ such that $\mathrm{P}U = u$ and $\mathrm{Q}U = \mng{u}$ are the first and second component of $U$, respectively.

\begin{definition}\label{def:el1}
First, we say that a lexicon $M$ contains an utterance $u$, i.e. $u \in_1 M$, if there is a UMP $U \in M$ such that $u = \mathrm{P}U$.
\end{definition}

\begin{definition}\label{def:el2}
Likewise, we say $M$ contains a meaning $\mng{u}$, i.e. $\mng{u} \in_2 M$ if there is a UMP $U \in M$ such that $\mng{u} = \mathrm{Q}U$.
\end{definition}

\begin{definition}\label{def:el3}
Furthermore, we say that $M$ contains an antecedent-consequent pair $(a, c)$, i.e.  $(a, c) \in_3 M$, if there is a UMP $U \in M$ such that $(a, c) \in \mathrm{Q}U$.
\end{definition}

Additionally, we define a \emph{local history} as a mapping from discrete time indices $t \in \mathbb{N}_0$ to UMPs.
\begin{definition}\label{def:hist}
A function $h: t \to U_t$, such that $h(t) = U_t$ is the UMP learned at time $t$, is called (local) history. At initialization $t = 0$ the history is given as a constant function
\begin{equation}\label{eq:inihist}
    h_0(t) = (\varepsilon, \emptyset)
\end{equation}
for all $t \in \mathbb{N}_0$.
During learning these entries are consecutively overwritten by the acquired UMPs $U_t$.
\end{definition}
The local histories store all UMP updates that happen during the learning process.

\begin{definition}\label{def:last}
Let $u \in_1 M$ be an utterance in the mental lexicon $M$. Then
\[
    \last(u) = \max \mng{u}
\]
is the latest update of $u$.
\end{definition}

After those preparations, we introduce our reinforcement algorithm as follows.

\begin{algorithm}\label{alg:reinf}

comprises the following rules that operate upon the current utterance $u_k$, its antecedent $a_k \in X$, and its consequent $c_k \in X$ at iteration $k$, and also at the mental lexicon $M_{k-1}$ depending on two crucial cases. The algorithm is initialized through $k = 1$, $t = 0$, $M_0 = \emptyset$, and the empty history \Eq{eq:inihist}.

\begin{enumerate}
  \item \label{it:r1} \emph{Case~1}. The utterance $u_k$ has never been uttered before, $u_k \not\in_1 M_{k-1}$.
  \begin{enumerate}
    \item \label{it:r1.1} \emph{Agree}. If the user's utterance is empty (silence), $u_k = \epsilon$, the system assumes that she agrees with the current operation mode. Hence, the system's state in $X$ does not change: $c_k = a_k$. Moreover, the lexicon is updated according to $M_k \gets M_{k-1} \cup \{ U_k \}$ with a new UMP $U_k = (u_k, \{(a_k, c_k) \})$.
        Subsequently, the local history is deleted
            \[
                h \gets h_0
            \]
            and reinitialized $t = 0$.
    \item \label{it:r1.2} \emph{Disagree}. If the user's utterance is not empty, $u_k \ne \epsilon$, the system assumes that the user does not agree with the current operation mode. Therefore, the system changes its epistemic state from the current one $a_k$ to another state, by drawing a conversational implicature $c_k = \Phi^\tau(a_k) \ne a_k$. A new UMP $U_k = (u_k, \{(a_k, c_k) \})$ is created and added to the lexicon through $M_k \gets M_{k-1} \cup \{ U_k \}$. Also $U_k$ is stored in the history: $h(t) \gets U_k$, $t \gets t + 1$.
  \end{enumerate}
  \item \label{it:r2} \emph{Case~2}. The utterance $u_k$ has been uttered before, $u_k \in_1 M_{k-1}$. Then, lexical access yields $U = (u, \mng{u}) \in M_{k-1}$ with $u = u_k$ as the current utterance, where we have to distinguish another two cases:
      \begin{enumerate}
        \item \label{it:r2.1} \emph{Apply}. There exists an antecedent-consequent pair $(a, c) \in \mng{u}$ such that the present state $a_k = a$. Then, apply the corresponding transition $a_k \mapsto c$ by means of a conventional implicature. If the antecedent $a_k$ has not accepted the utterance $u$ (i.e. $a_k \ne c_k$), a new UMP $U = (u, \{(a, c) \})$ is created to be stored in the history $h(t) \gets U$, $t \gets t + 1$ afterwards. If, contrastingly, the antecedent $a_k$ has accepted the utterance $u$ (i.e. $a_k = c_k$), then the local history is deleted
            \[
                h \gets h_0
            \]
            and reinitialized $t = 0$. Finally, exchange the ordering of the applied antecedent-consequent pair and the last one in $\mng{u}$ to allow for subsequent revision, hence
            \[
                \mng{u} \gets (\mng{u} \setminus \{(a, c) \}) \cup \{(a, c) \} \:.
            \]
        \item \label{it:r2.2} \emph{Learn}. There is no antecedent-consequent pair $(a, c) \in \mng{u}$ such that $a_k = a$. In this case, the agent has to augment the meaning of $\mng{u}$ as follows.
        \begin{enumerate}
            \item \label{it:r2.2.1} \emph{Agree}. If the user's utterance is empty (silence), $u_k = \epsilon$, the system assumes that the user endorses the current operation mode. Therefore $c_k = a_k$, again. Then, firstly, $M_k \gets M_{k-1} \setminus \{U \}$, secondly, $U$ is updated according to $\mng{\epsilon} \gets \mng{\epsilon} \cup \{ U_k \}$ with a newly created $U_k = (a_k, c_k)$, thirdly, $M_k \gets M_k \cup \{U \}$, fourthly, delete and reinitialize the local history
                \[
                    h \gets h_0 \:, \quad t = 0 \:.
                \]
            \item \label{it:r2.2.2} \emph{Disagree}. If the user's utterance is not empty, $u_k \ne \epsilon$, the system assumes that the user does not agree with the current operation mode. Thus, it makes a state transition
                through a conversational implicature $c_k = \Phi^\tau(a_k) \ne a_k$.
                Then, again firstly, $M_k \gets M_{k-1} \setminus \{U \}$, secondly, $U$ is updated according to $\mng{u} \gets \mng{u} \cup \{ U_k \}$ with $U_k = (a_k, c_k)$, thirdly,  $M_k \gets M_k \cup \{U \}$, and fourthly, store $U_k$ in the history: $h(t) \gets U_k$, $t \gets t + 1$.
        \end{enumerate}
      \end{enumerate}
      \item \label{it:r3} \emph{Revise}. If an utterance $u_k$, that is not accepted by the current antecedent $a_k$ (i.e. $\mng{u_k}(a_k) \ne a_k$) immediately follows a history of unaccepted utterances, then the recently acquired meaning of $u_k$ revises that of all earlier utterances in the current local history as follows. For all times $s < t$, the latest update $(a_s, c_s) = \last(u_s)$ of the utterances $u_s = U_s(1)$ with $U_s = h(s)$ is retrieved and revised according to
           \[
                \mng{u_s} \gets (\mng{u_s} \setminus \{(a_s, c_s) \}) \cup \{(a_s, c_k) \}
           \]
        with $c_k \ne c_s$. Afterwards the revised $U_s$ is stored in the history: $h(t) \gets U_s$, $t \gets t + 1$.
\end{enumerate}
\end{algorithm}

Algorithm \ref{alg:reinf} continuously checks whether an utterance has been presented to the agent before. If this was not the case, rule \ref{it:r1} applies, otherwise, the system switches into mode \ref{it:r2}. In both cases, we assume that an empty utterance (\ref{it:r1.1}, \ref{it:r2.2.1}) means that the operator endorses the current action state of the machine. However, if the utterance is not empty, the user does not agree with the new machine state (\ref{it:r1.2}, \ref{it:r2.2.2}). For our simple model of the cognitive heating with binary action space, \Eq{eq:actions}, the current state $a$ is just changed into its opposite state $\neg a$. Yet, for a more complex setting with more than two states, the agent has to find the appropriate response. The most simple way to achieve this, is by probing all possible consequents through trial and error. More suitable solutions could exploit Markov decision processes (MDP) for instance \citep{SuttonBarto18}. Particularly special cases are the rules \ref{it:r2.1} and \ref{it:r3}. Apply \ref{it:r2.1} takes place when the utterance belongs to the acquired lexicon and the antecedent of its meaning is associated with a known consequent, expressing a conventional implicature made by the machine. Then, the agent simply carries out the prescribed state transition. Finally, our revision rule \ref{it:r3} revises an entire history of unaccepted utterances immediately following the most recently accepted utterance that reinitialized a local history. The revision rule rests on the implicit assumption (another kind of conversational implicature) that the state transition induced by the current utterance is the one actually \emph{meant} by the user, thereby replacing all ``wrongly'' learned consequents in the history by the current one, such that all  previously learned antecedents are mapped to the current consequent. Moreover, this rule assumes that a sequence of unaccepted utterances indicates an abiding misunderstanding between the user and the agent that must be ruled out at once by the latest revision.

One obvious property of the algorithm \ref{alg:reinf} is stated as the following proposition.
\begin{proposition}\label{prop:affi}
The meaning of silence accepts any epistemic state, $\mng{\epsilon} = \id$.
\end{proposition}

\noindent\emph{Proof}. $\mng{\epsilon}(x) = \id(x) = x$ for all $x \in X$.

Next, we have to consider whether the proposed algorithm is \emph{sound} and \emph{complete}. Clearly, the algorithm cannot be complete for it attributes meaning to principally any utterance, even to interjections such as ``uh'', ``eh'', and ``mmh''. By contrast, we may regard soundness as the constraint that the acquired relations $F$ between antecedent-consequent pairs is a proper partial function in the sense of \Eq{eq:mean}. Then we could prove soundness as follows.

\begin{theorem}\label{theo:sound}
The learning algorithm \ref{alg:reinf} is sound.
\end{theorem}

\noindent\emph{Proof}. Assume that the algorithm has already learned that the meaning of an utterance $u$ contains one antecedent-consequent pair $(a, c) \in \mng{u}$. Further assuming that another pair $(a, c')$, $c' \ne c$ should be learned as well. This does not happen, because the algorithm recognizes the antecedent $a$ and goes into \emph{apply} mode according to rule \ref{it:r2.1} instead of learning the meaning $(a, c') \in \mng{u}$ which would yield an inconsistent partial function $F$.

Finally, we give some indication of a notion of convergence.
\begin{definition}\label{def:conver}
The reinforcement algorithm is said to \emph{converge} if it learns in reasonable time without learning too many utterances and without too many revisions.
\end{definition}
Clearly, definition \ref{def:conver} is rather problematic without a precisely prescribed topology in training space. However, even this provisional definition leads to the following main result of our approach.

\begin{theorem}\label{theo:convergence}
The learning algorithm \ref{alg:reinf} converges when the user is committed to obey the Cooperative Principle and the Conversation Maxims given in Tables \ref{tab:grice}, \ref{tab:horn}, and their respective formalizations (definitions  \ref{def:bq}, \ref{def:br}, \ref{def:q1}, \ref{def:q2}).
\end{theorem}

\noindent\emph{Proof}. We check the consequences of a user who is voluntarily violating the Principles Q, R and Quality 1, and 2. We always stipulate that the machine is in an antecedent state $a \in X$ and the user desires a transition into consequent state $c$ by uttering $u$.

First, assume that the user violates Principle Q. In this case, she has to decide between two equally complex utterances $u_1, u_2$ that may drastically differ in terms of their informativity. Suppose $\mng{u_1} \subset \mng{u_2}$ such that $\mng{u_2}$ is much more informative than $\mng{u_1}$. Then Principle Q advises her to utter $\mng{u_2}$ by maximizing the hearers informativity. In that case, it is much more likely, that the desired state transition $(a, c) \in \mng{u_2}$, than $(a, c) \in \mng{u_1}$. Thus, uttering $u_2$ leads to a conventional implicature according to rule \ref{it:r2.1} with higher probability than uttering $u_1$, which in turn could entail a conversational implicature, endangering subsequent revisions after rule \ref{it:r3}.

Next, we consider a user violating Principle R. This seems to be less problematic at a first glance as the user acts adversely against her own interest of minimizing production effort. Thus, it seems much more rational when the user addresses the machine by brief imperatives such as ``heat!'', instead of uttering ``I am going to grandma'', although the conversational implicatures  drawn by the machine are essentially the same. Even worse would be another case of Principle R violation, namely a user who permanently produces verboseness without any regular repetition. Then, the algorithm is captured by the learning mode \ref{it:r1.2} and will rarely escape.

Regarding Principle Quality 1, we have already seen that a user who desires the state transition $a \mapsto c$ but who produces knowingly and purposely utterances $u'$ that cannot entail the desired transition because $(a, c) \not\in \cg(u')_{s,h} \ne \emptyset $, says something she knows to be false. As such utterances will never lead to an accepting state, they cause repetitive revisions (rule \ref{it:r3}), thereby preventing convergence.

Almost the same happens under violation of Principle Quality 2, where the only difference is that the user definitely does know that a desired transition $a \mapsto c$ is prohibited by uttering $u'$ because such a transition is simply not defined, for $(a, c) \not\in \cg(u')_{s,h} = \emptyset $. Thus, her utterance would lack sufficient evidence. In this case, the algorithm is captured by mode \ref{it:r1.2} and does not escape.

\section{Results}
\label{sec:results}

In this section we present two different training scenarios for our example of a cognitive heating device \citep{KlimczakWolffLindemannEA14, HuberEA18}. In the first scenario, we discuss the motivating example from \Sec{sec:dynsem} with only two operation modes: heating ($H$) and non-heating ($\neg H$) [\Eq{eq:actions}]. The second scenario then deals with a slightly more complex umwelt comprising three actions: non-heating ($\neg H$), semi-heating $S$ and full-heating $H$. We initialize the system at time $k = 0$ with empty lexicon $M_0 = \emptyset$ and its current operation mode $H$ (assuming that the heating heats at time $k = 0$). Therefore the antecedent state $a_0 = H$. To the behavior $b_0$ we associate the given utterance $u_0$.

\subsection{First Scenario}
\label{sec:1scen}

In the first scenario we train the system with the following sequence of utterances:
\begin{multline}\label{eq:1scut}
  \mathbf{s}_1 = (\epsilon, \TT{I am going to grandma}, \epsilon, \TT{I am going to grandma}, \\
    \TT{no!}, \epsilon, \TT{heat!}, \TT{I am going to grandma}, \\
    \TT{I am going to grandma}, \epsilon, \TT{good boy!}, \\
    \TT{I am going to grandma}, \TT{I am going to grandma} ) \:.
\end{multline}
We use typewriter font to denote the transcription of the speech tokens, thereby abstracting from its conventional and compositional meaning in the first place.

According to the intuition behind algorithm \ref{alg:reinf}, the user approves the initial working state by an empty utterance. Then, she changes her mind and desires a state transition by uttering \TT{I am going to grandma}, which is subsequently endorsed by silence in the third iteration. By repeating the last utterance in the fourth step, another state change is requested. However, the corresponding transition is punished by the utterance \TT{no!} and the meaning of \TT{I am going to grandma} becomes revised through reinforcement learning. In the sixth iteration, the current operation mode is endorsed by the users silence. In iteration seven, we demonstrate that the machine is even able to cope with confounding inputs, when a confused user firstly utters \TT{heat!} and secondly corrects herself by \TT{I am going to grandma}, desiring the heating to be turned off. Then, the user appears as completely deranged when repeating the last utterance again. The correction is rewarded by the empty utterance in step ten. Finally, we show how the algorithm can also learn the partially affirmative meaning of \TT{good boy!} in three further reinforcement steps.

\paragraph{First iteration.}
\label{sec:1iter1}

For the sake of simplicity, we assume that the user remains silent at time $k = 0$, such that the initial utterance is the empty word $u_0 = \epsilon$. Then,
\[
	u_0 = \epsilon
\]
at time $k= 0$. Thus, rule \ref{it:r1.1} applies and the system does not change its operation mode while the user remains silent. As a consequence we obtain the epistemic state
\[
	c_0 = H \:.
\]
Thereby the first meaning becomes
\[
	\mng{\epsilon} = \{ (H, H) \} \:.
\]
Correspondingly, the learned UMP is
\[
	U_0 = (\epsilon, \{ (H, H) \}) \:.
\]

According to rule \ref{it:r1.1} the lexicon gets
\[
	M_1 = \{ (\epsilon, \{ (H, H) \} ) \} \:.
\]

\paragraph{Second iteration.}
\label{sec:2iter1}

At the next iteration at time $k=1$, the antecedent becomes dynamically initialized with the consequent from the previous time step [\Eq{eq:iniup}]:
\[
	a_1 \gets c_0 \:.
\]	
In our particular case, we have $a_1 = H$. As before, we associate to the current verbal behavior $b_1$ the utterance $u_1$, now letting  $u_1 = \TT{I am going to grandma}$. Then,
\[
	u_1 = \TT{I am going to grandma} \:,
\]
and
\[
	c_1 = \neg H
\]
according to training rule \ref{it:r1.2} where a non-empty utterance indicates the users' desire to change the heating's operation mode from heating $H$ to non-heating $\neg H$. Therefore,
\[
	\mng{\TT{I am going to grandma}} = \{ (H, \neg H) \} \:.
\]
The resulting UMP is then
\[
	U_1 = (\TT{I am going to grandma}, \{ (H, \neg H) \} ) \:.
\]

Since the UMP $U_1$ is not contained in the lexicon $M_1$ from the first iteration, we simply obtain the updated mental lexicon
\[
  M_2 = M_1 \cup \{ U_1 \} \:,
\]
i.e.
\[
	M_2 = \{ (\epsilon, \{ (H, H) \} ), (\TT{I am going to grandma}, \{ (H, \neg H) \} ) \} \:.
\]

\paragraph{Third iteration.}
\label{sec:3iter1}

The third iteration starts with antecedent $a_2 = c_1 = \neg H$ and processes utterance $u_2 = \epsilon$ due to \eqref{eq:1scut}. Since the utterance is already known to the device, it has to consider rules \ref{it:r2}. As it does not find the antecedent $\neg H$ in the meaning definition $\mng{\epsilon}$, it proceeds with rule \ref{it:r2.2.1}. Thus,
\begin{eqnarray*}
    U &=& (\epsilon, \{ (H, H) \}) \\
	U_2 &=& (\epsilon, \{ (\neg H, \neg H) \} ) \\
    \mng{\epsilon} &\gets& \mng{\epsilon} \cup \{ (\neg H, \neg H) \} \\
    U &\gets& (\epsilon, \{ (H, H),  (\neg H, \neg H)\}) \\
    M_3 &=& \{ (\epsilon, \{ (H, H), (\neg H, \neg H) \} ), (\TT{I am going to grandma}, \{ (H, \neg H) \} ) \} \:.
\end{eqnarray*}

\paragraph{Fourth iteration.}
\label{sec:4iter1}

In the next iteration we have antecedent $a_3 = c_2 = \neg H$ and interpret utterance $u_3 = \TT{I am going to grandma}$ again [\Eq{eq:1scut}]. The utterance is known to the agent, and we adopt rule \ref{it:r2.2.2} because the antecedent does not occur in the previous meaning definition. Then,
\begin{eqnarray*}
    U &=& (\TT{I am going to grandma}, \{ (H, \neg H) \}) \\
	U_3 &=& (\TT{I am going to grandma}, \{ (\neg H, H)^* \} ) \\
    \mng{\TT{I am going to grandma}} &\gets& \mng{\TT{I am going to grandma}} \cup \{ (\neg H, H)^* \} \\
    U &\gets& (\TT{I am going to grandma}, \{ (H, \neg H),  (\neg H, H)^*\}) \:.
\end{eqnarray*}

Note that the newly acquired meaning $(\neg H, H) \in \mng{\TT{I am going to grandma}}$ is pragmatically not appropriate \citep{Grice89}. We henceforth indicate this kind of ``wrong'' knowledge by the asterisk ``$*$''.

Thereby, the updated lexicon becomes
\begin{multline*}
    M_4 = \{ (\epsilon, \{ (H, H), (\neg H, \neg H) \} ), (\TT{I am going to grandma}, \{ (H, \neg H), (\neg H, H)^* \} ) \} \:.
\end{multline*}

\paragraph{Fifth iteration.}
\label{sec:5iter1}

The meaning of \TT{I am going to grandma} acquired in the last training step is at variance with the pragmatic implicature discussed in \Sec{sec:intro} because the heating is turned on when it was off previously \citep{Grice89}. Therefore, the user utters \TT{no!} First of all, as $u_4 \ne \epsilon$, the user does not consent with the antecedent state $a_4 = H$ and desires consequent $c_4 = \neg H$, instead. Hence, the agent learns a new meaning
\[
	\mng{\TT{no!}} = \{ (H, \neg H) \} \:.
\]
The learned UMP is then
\[
	U_4 = (\TT{no!}, \{ (H, \neg H) \}) \:.
\]

According to rule \ref{it:r1.1} the lexicon gets
\[
	M_5 = M_4 \cup \{ (\TT{no!}, \{ (H, \neg H) \}) \} \:.
\]

However, as $u_4$ immediately follows the utterance $u_3$, we also have to take the revision rule \ref{it:r3} of the reinforcement learning algorithm into account. To this aim, the machine checks whether the antecedent $a_4 = H$ was accepted by the meaning of $u_4$. This was not the case, as
\[
    c_4 = \mng{\TT{no!}}(a_4) = \mng{\TT{no!}}(H) = \neg H \:.
\]
Hence, the consequent in the already learned meaning definition of $u_3$ is replaced by the currently desired consequent. Thus,
\[
    \mng{\TT{I am going to grandma}} \gets (\mng{\TT{I am going to grandma}} \setminus \{( \neg H, H ) \}) \cup \{(\neg H, \neg H) \}
\]
and thus
\begin{multline*}
    M_5 \gets \{ (\epsilon, \{ (H, H), (\neg H, \neg H) \} ), (\TT{I am going to grandma}, \{ (H, \neg H), (\neg H, \neg H) \} ) , (\TT{no!}, \{ (H, \neg H) \} ) \} \:.
\end{multline*}

As a consequence, \TT{I am going to grandma} assumes the intended meaning in the lexicon $M_5$, i.e.
\begin{equation}\label{eq:igtgn}
  \mng{\TT{I am going to grandma}} = \{ (H, \neg H), (\neg H, \neg H) \} \:,
\end{equation}
i.e. turn off the heating anyways, which is the correct pragmatic implicature that no heating power is required in case of the user's absence.

\paragraph{Sixth iteration.}
\label{sec:6iter1}

In the next iteration, where $a_5 = c_4 = \neg H$, the user expresses consent with the current working state of the heating by empty utterance $u_5 = \varepsilon$. Because $u_5$ has been uttered before and its antecedent $a_5$ is known in the lexicon, rule \ref{it:r2.1}, i.e, \emph{apply} takes place: the working state does not change, there is no learning and nothing is revised.

So far all rules of algorithm \ref{alg:reinf} have been applied. As stated in proposition \ref{prop:affi}, the empty utterance $\varepsilon$ is accepted by all epistemic states, which can be interpreted as an affirmation. The meaning of \TT{I am going to grandma} has been learned as the intended pragmatic implicature that no heating power is required during absence of the user. The meaning of \TT{no!} has been partially acquired. Finally, the revision rule \ref{it:r3} was triggered by an unaccepted utterance \TT{no!}.

\paragraph{Seventh iteration.}
\label{sec:7iter1}

Next, we investigate how the cognitive heating is able to cope with confounding information uttered by a slightly bemused user. Therefore, the machine has to process $u_6 = \TT{heat!}$ in state $a_6 = c_5 = \neg H$. For $u_6$, which is not empty, has never been encountered before, rule \ref{it:r1.2} applies as
\[
	\mng{\TT{heat!}} = \{ (\neg H, H) \} \:.
\]
The resulting UMP is then
\[
	U_6 = (\TT{heat!}, \{ (\neg H, H) \} ) \:.
\]

Because the UMP $U_6$ is not contained in the lexicon $M_5$ from the last iteration, we obtain the updated mental lexicon
\[
  M_6 = M_5 \cup \{ U_6 \} \:,
\]
i.e.
\begin{multline*}
	M_6 = \{ (\epsilon, \{ (H, H), (\neg H, \neg H) \} ), (\TT{I am going to grandma}, \{ (H, \neg H), (\neg H, \neg H) \} ) , \\
    (\TT{no!}, \{ (H, \neg H) \}) ,  (\TT{heat!}, \{ (\neg H, H) \} ) \} \:.
\end{multline*}

\paragraph{Eighth iteration.}
\label{sec:8iter1}

Although the last utterance has adopted the correct meaning, it is pragmatically inappropriate. Thus, the user makes a correction by uttering $u_7 = \TT{I am going to grandma}$, again. First, the dynamic meaning of $u_7$ is applied to the current state $a_7 = c_6 = H$ according to rule \ref{it:r2.1}, leading to the state transition
\[
    c_7 = \mng{\TT{I am going to grandma}}(a_7) = \mng{\TT{I am going to grandma}}(H) = \neg H \:.
\]

Second, since $c_7 \ne a_7$, the antecedent $a_7$ does not accept the meaning of \TT{I am going to grandma}, triggering thereby a revision of $u_6$, which leads to
\[
    \mng{\TT{heat!}} \gets (\mng{\TT{heat!}} \setminus \{ (\neg H, H) \} ) \cup \{ (\neg H, \neg H)^* \}
\]
where the learned antecedent is (erroneously) associated with the current consequent.

\paragraph{Nineth iteration.}
\label{sec:9iter1}

Now we assume that the user appears deranged to such an extent that she repeats the last utterance once again: $u_8 = \TT{I am going to grandma}$. Because the working state is $a_8 = c_7 = \neg H$ which belongs to the meaning definition of $u_8$, this is applied through
\[
    c_8 = \mng{\TT{I am going to grandma}}(a_8) = \mng{\TT{I am going to grandma}}(\neg H) = \neg H = a_8
\]
which is a fixed point of $\mng{\TT{I am going to grandma}}$. Therefore, the antecedent $a_8$ accepts the meaning of $u_8$, preventing further revision.

\paragraph{Tenth iteration.}
\label{sec:10iter1}

The user gives silent consent. Here we have $a_9 = c_8 = \neg H$ and $u_9 = \varepsilon$, which is straightforwardly applied as
\[
    c_9 = \mng{\varepsilon}(a_9) = \mng{\varepsilon}(\neg H) = \neg H = a_9 \:.
\]

\paragraph{Eleventh iteration.}
\label{sec:11iter1}

We finish the first scenario by the acquisition of affirmative meanings. In state $a_{10} = c_9 = \neg H$, the user utters $u_{10} = \TT{good boy!}$ which is a non-empty unknown phrase that leads to lexicon update according to rule \ref{it:r1.2}, such that
\[
	\mng{\TT{good boy!}} = \{ (\neg H, H)^* \} \:.
\]
The resulting UMP is then
\[
	U_{10} = (\TT{good boy!}, \{ (\neg H, H)^*  \} ) \:.
\]

Since UMP $U_{10}$ is not contained in the current lexicon this becomes updated through
\begin{multline*}
	M_7 = \{ (\epsilon, \{ (H, H), (\neg H, \neg H) \} ), (\TT{I am going to grandma}, \{ (H, \neg H), (\neg H, \neg H) \} ) , \\
    (\TT{no!}, \{ (H, \neg H) \}) ,  (\TT{heat!}, \{ (\neg H, H) \} ),  (\TT{good boy!}, \{ (\neg H, H)^*  \} ) \} \:.
\end{multline*}

\paragraph{Twelfth iteration.}
\label{sec:12iter1}

Because, the meaning of \TT{good boy!} has been erroneously acquired in the last iteration, the user immediately utters $u_{11} = \TT{I am going to grandma}$ to enforce a revision in state $a_{11} = c_{10} = H$. Its application yields
\[
    c_{11} = \mng{\TT{I am going to grandma}}(a_{11}) = \mng{\TT{I am going to grandma}}(H) = \neg H \:.
\]

Hence, the meaning of \TT{I am going to grandma} is not accepted by the antecedent $a_{11}$, thereby causing $u_{10}$ to be revised, leading to
\[
    \mng{\TT{good boy!}} \gets ( \mng{\TT{good boy!}} \setminus \{ (\neg H, H) \} ) \cup \{ (\neg H, \neg H) \} \:.
\]
Now, the meaning of \TT{good boy!} accepts the state $\neg H$ as desired.

\paragraph{Thirteenth iteration.}
\label{sec:13iter1}

The still slightly confused user terminates this scenario by another reassuring $u_{12} = \TT{I am going to grandma}$ in operation mode $a_{12} = c_{11} = \neg H$ which leads to a final application of rule \ref{it:r2.1}:
\[
    c_{12} = \mng{\TT{I am going to grandma}}(a_{12}) = \mng{\TT{I am going to grandma}}(\neg H) = \neg H = a_{12}
\]
accepting the antecedent $a_{12}$. Therefore, no revision is induced.

Finally, we condense the epistemic updating dynamics of this scenario in \Tab{tab:sc1}.

\begin{table}[H]
\centering 
\caption{\label{tab:sc1} Dynamic Machine Semiotics for the first scenario \eqref{eq:1scut}. The asterisk$^*$ indicates wrong knowledge. Variables are $k$: iteration, $t$: local time stamp, $a_k$: antecedent, utterance $u_k$ (i.e. behavior $b_k$), $c_k$: consequent, $U_t$: learned utterance-meaning pair (UMP), rule: the applied rule of algorithm \ref{alg:reinf}. The history from definition \ref{def:hist} is rendered from the penultimate column.
}
\begin{tabular}{rrllllr}
  \hline
  $k$ & $t$ & $a_k$ & $u_k$ & $c_k$ & $U_t$  &  rule \\
  \hline
    0 & 0 & $H$ &	$\epsilon$ &			$H$ &		$(\epsilon, \{ (H, H) \})$ & \ref{it:r1.1} \\
    1 & 0	 & $H$ &	\TT{I am going to grandma} &	$\neg H$ &		$(\TT{I am going to grandma}, \{ (H, \neg H) \} )$ & \ref{it:r1.2} \\
    2 & 1 & $\neg H$ &	$\epsilon$ &			$\neg H$ &		$(\epsilon, \{ (\neg H, \neg H)\})$ & \ref{it:r2.2.1} \\
    3 & 0 & $\neg H$ &	\TT{I am going to grandma} &	$H$ &		$(\TT{I am going to grandma}, \{ (\neg H, H)^*  \})$ & \ref{it:r2.2.2} \\
    4	& 1	& $H$ &	\TT{no!} &	$\neg H$ &		$(\TT{no!}, \{ (H, \neg H) \})$ & \ref{it:r1.2} \\
    & 2  &  &	 &	 &		$(\TT{I am going to grandma}, \{ (\neg H, \neg H) \})$ & \ref{it:r3} \\
    5	& 3  & $\neg H$ & $\epsilon$  &	$\neg H$ &	$(\epsilon, \{ (\neg H, \neg H)\})$	& \ref{it:r2.1} \\
    6 & 0 	& $\neg H$ &	\TT{heat!} &	 $H$ &		$(\TT{heat!}, \{ (\neg H, H) \})$ & \ref{it:r1.2} \\
    7 & 1 	& $H$ &	\TT{I am going to grandma} &	$\neg H$ &		$(\TT{I am going to grandma}, \{ (H, \neg H) \} )$ & \ref{it:r2.1} \\
    & 2	&  &	 &	 &		$(\TT{heat!}, \{ (\neg H, \neg H)^* \})$ & \ref{it:r3} \\
    8 & 3 	& $\neg H$ &	\TT{I am going to grandma} &	$\neg H$ &		$(\TT{I am going to grandma}, \{ (\neg H,\neg H)  \})$ & \ref{it:r2.1} \\
    9 & 0  & $\neg H$ & $\epsilon$  &	$\neg H$ &	$(\epsilon, \{ (\neg H, \neg H)\})$	& \ref{it:r2.1} \\
    10 & 0 	& $\neg H$ &	$\TT{good boy!}$ &			$H$ &		$(\TT{good boy!}, \{ (\neg H,  H)^* \})$ & \ref{it:r1.2} \\
    11 & 1	& $H$ &	\TT{I am going to grandma} &	$\neg H$ &		$(\TT{I am going to grandma}, \{ (H, \neg H) \} )$ & \ref{it:r2.1} \\
    & 2 & & & &		$(\TT{good boy!}, \{ (\neg H,  \neg H) \})$   & \ref{it:r3} \\
    12 & 3 & $\neg H$ &	\TT{I am going to grandma} &	$\neg H$ &		$(\TT{I am going to grandma}, \{ (\neg H,\neg H)  \})$ & \ref{it:r2.1} \\
  \hline
\end{tabular}
\end{table}

\subsection{Second Scenario}
\label{sec:2scen}

In the second scenario we consider an augmented epistemic space $X$ comprising three operation modes: don't heat $\neg H$, heat $H$ and semi-heat $S$. We also consider a more complex intrinsic state dynamics $\Phi$ that circulates between these modes through $\Phi(\neg H) = S$, $\Phi(S) = H$, and $\Phi(H) = \neg H$, for all unknown utterances (cf. \Sec{sec:dynsem}). This intrinsic dynamics is clearly ergodic.

Due to the different umwelt, we process a slightly deviant sequence of utterances here:
\begin{multline}\label{eq:2scut}
  \mathbf{s}_2 = (\epsilon, \TT{I am going to grandma}, \epsilon, \TT{I am going to grandma}, \\
    \TT{no!}, \TT{no!}, \epsilon, \TT{heat!}, \TT{I am going to grandma}, \TT{I am going to grandma}, \epsilon, \TT{good boy!}, \\
    \TT{I am going to grandma}, \TT{I am going to grandma} ) \:.
\end{multline}
The only difference in comparison to scenario \eqref{eq:1scut} is the repetition of \TT{no!} in step six.

The first ten steps of the resulting updating dynamics of this scenario are depicted in \Tab{tab:sc2a}.

\begin{table}[H]
\centering 
\caption{\label{tab:sc2a} Beginning of the Machine Semiotics for the second scenario \eqref{eq:1scut}. The asterisk$^*$ indicates wrong knowledge. Variables are $k$: iteration, $t$: local time stamp, $a_k$: antecedent, utterance $u_k$ (i.e. behavior $b_k$), $c_k$: consequent, $U_t$: learned utterance-meaning pair (UMP), rule: the applied rule of algorithm \ref{alg:reinf}. The history from definition \ref{def:hist} is rendered from the penultimate column.}
\begin{tabular}{rrllllr}
  \hline
  $k$ & $t$ & $a_k$ & $u_k$ & $c_k$ & $U_t$  &  rule \\
  \hline
    0 & 0 	& $H$ &	$\epsilon$ &			$H$ &		$(\epsilon, \{ (H, H) \})$ & \ref{it:r1.1} \\
    1 & 0 	& $H$ &	\TT{I am going to grandma} &	$\neg H$ &		$(\TT{I am going to grandma}, \{ (H, \neg H) \} )$ & \ref{it:r1.2} \\
    2 & 1 	& $\neg H$ &	$\epsilon$ &			$\neg H$ &		$(\epsilon, \{ (\neg H, \neg H)\})$ & \ref{it:r2.2.1} \\
    3 & 0 	& $\neg H$ &	\TT{I am going to grandma} &	$S$ &		$(\TT{I am going to grandma}, \{ (\neg H, S)^*  \})$ & \ref{it:r2.2.2} \\
    4 & 1 	& $S$ &	\TT{no!} &	$H$ &		$(\TT{no!}, \{ (S, H) \})$ & \ref{it:r1.2} \\
    & 2 & &  &	 &	 $(\TT{I am going to grandma}, \{ (\neg H, H)^* \})$ & \ref{it:r3} \\
    5 & 3 	& $H$ &	\TT{no!} &	$\neg H$ &		$(\TT{no!}, \{ (H, \neg H) \})$ & \ref{it:r2.2.2} \\
    & 4  &  &	 &	 &		$(\TT{no!}, \{ (S, \neg H) \})$ & \ref{it:r3} \\
    & 5  &  &	 &	 &		$(\TT{I am going to grandma}, \{ (\neg H, \neg H) \})$ & \ref{it:r3} \\
    6 & 6  & $\neg H$ & $\epsilon$  &	$\neg H$ &	$(\epsilon, \{ (\neg H, \neg H)\})$	& \ref{it:r2.1} \\
  \hline
\end{tabular}
\end{table}

Steps 0 to 2 are the very same as in the first scenario shown in \Tab{tab:sc1}. Yet note that the empty utterance $\epsilon$ was accepted by state $\neg H$ in step 2. Beginning with step 3 the Machine Semiotics evolves differently due to the presence of three action states, when the meaning of  \TT{I am going to grandma} is erroneously acquired as $(\neg H, S)$. In step 4, the user therefore triggers a first revision by uttering \TT{no!} that is, though pragmatically undesired, carried out at local time $t = 2$. Therefore, the user repeats \TT{no!} in iteration 5. At this point the user-machine communication has accumulated abiding misunderstandings which must be resolved at once in the succeeding steps. This requires the formal concepts prepared in \Sec{sec:rforlear}.

First the system renders the history $h$ that is scanned for local time indices $s = 1, \dots, 5$ for the last meaning updates. At local time $t = 4$ the meaning of \TT{no!} learned at iteration 4 is revised by replacing the acquired consequent $H$ by the newly desired consequent $\neg H$ while keeping its antecedent $S$. Moreover, at time $t = 5$ the pragmatically inappropriate meaning of \TT{I am going to grandma} is revised from $(\neg H, H)$ to $(\neg H, \neg H)$, i.e. ``don't heat during absence of the user''. The first training phase terminates at time 6 by accepting the empty utterance in state $\neg H$, thereby deleting and reinitalizing the local history (not shown).

The mental lexicon acquired sofar is given as
\begin{multline*}
	M_7 = \{ (\epsilon, \{ (H, H), (\neg H, \neg H) \} ), (\TT{I am going to grandma}, \{ (H, \neg H), (\neg H, \neg H) \} ) , \\
    (\TT{no!}, \{ (H, \neg H), (S, \neg H) \}) \} \:.
\end{multline*}
In $M_7$ the empty utterance $\epsilon$ acts as a partial affirmation accepting any state. The utterance \TT{I am going to grandma} pragmatically implicates that no heating power is required when the user is departed. Finally, \TT{no!} means that neither full nor semi-heating are desired.

The next Table \ref{tab:sc2b} presents the continuation of the three-state Machine Semiotics in case of confounding information from a somewhat bewildered user.

\begin{table}[H]
\centering 
\caption{\label{tab:sc2b} Continuation of the Machine Semiotics for the second scenario \eqref{eq:1scut}. The asterisk$^*$ indicates wrong knowledge. Variables are $k$: iteration, $t$: local time stamp, $a_k$: antecedent, utterance $u_k$ (i.e. behavior $b_k$), $c_k$: consequent, $U_t$: learned utterance-meaning pair (UMP), rule: the applied rule of algorithm \ref{alg:reinf}. The history from definition \ref{def:hist} is rendered from the penultimate column.
}
\begin{tabular}{rrllllr}
  \hline
  $k$ & $t$ & $a_k$ & $u_k$ & $c_k$ & $U_t$  &  rule \\
  \hline
   7 & 0 & $\neg H$ &	\TT{heat!} &	 $S$ &		$(\TT{heat!}, \{ (\neg H, S) \})$ & \ref{it:r1.2} \\
   8 & 1	 & $S$ &	\TT{I am going to grandma} &	$H$ &		$(\TT{I am going to grandma}, \{ (S, H)^* \} )$ & \ref{it:r2.2.2} \\
   & 2 &  &	 &	 &		$(\TT{heat!}, \{ (\neg H, H) \})$ & \ref{it:r3} \\
   9 & 3	 & $H$ &	\TT{I am going to grandma} &	$\neg H$ &		$(\TT{I am going to grandma}, \{ (H, \neg H) \} )$ & \ref{it:r2.1} \\
   &  4 &  &	 &	 &		$(\TT{I am going to grandma}, \{ (S, \neg H) \})$ & \ref{it:r3} \\
   &  5 &  &	 &	 &		$(\TT{heat!}, \{ (\neg H, \neg H)^* \})$ & \ref{it:r3} \\
  \hline
\end{tabular}
\end{table}

Although the user wants to visit her grandmother, she utters \TT{heat!} which would have been correctly learned as switching into (semi-)heating mode from non-heating under normal circumstances. However, as heating power is unwanted in the present context, she says \TT{I am going to grandma} at iteration 8 again. This triggers another revision where the previously associated consequent $S$ is replaced by the current $H$ in the meaning of \TT{heat!}. Thereby, the machine even further increases its heating power, instead of shutting down. Hence the user repeats the correctly learned \TT{I am going to grandma} at iteration 9.

Now the system has again accumulated abiding misunderstanding that must be corrected at once. To this end, the current local history $h$ is scanned for the last updates. As a consequence, the meanings of \TT{I am going to grandma} and \TT{heat!} are revised according to the rule \ref{it:r3} of algorithm \ref{alg:reinf}.

The mental lexicon acquired sofar is then
\begin{multline*}
	M_{10} = \{ (\epsilon, \{ (H, H), (\neg H, \neg H) \} ), \\
    (\TT{I am going to grandma}, \{ (H, \neg H), (\neg H, \neg H) , (S, \neg H) \} ) , \\
    (\TT{no!}, \{ (H, \neg H), (S, \neg H) \}), (\TT{heat!}, \{ (\neg H, \neg H)^* \}) \} \:,
\end{multline*}
where the meaning of \TT{I am going to grandma} has been correctly augmented by another antecedent-consequent pair $(S, \neg H)$, while the meaning of \TT{heat!} became ``don't heat'' when the heater was turned off before, which is at variance with the intended implicature.

We conclude this scenario with the acquisition of a corroboration meaning again. This is shown in \Tab{tab:sc2c}.

\begin{table}[H]
\centering 
\caption{\label{tab:sc2c} Finalization of the Machine Semiotics for the second scenario \eqref{eq:1scut}. The asterisk$^*$ indicates wrong knowledge. Variables are $k$: iteration, $t$: local time stamp, $a_k$: antecedent, utterance $u_k$ (i.e. behavior $b_k$), $c_k$: consequent, $U_t$: learned utterance-meaning pair (UMP), rule: the applied rule of algorithm \ref{alg:reinf}. The history from definition \ref{def:hist} is rendered from the penultimate column.
}
\begin{tabular}{rrllllr}
  \hline
  $k$ & $t$ & $a_k$ & $u_k$ & $c_k$ & $U_t$  &  rule \\
  \hline
    10 & 6 & $\neg H$ &	$\epsilon$ &			$\neg H$ &		$(\epsilon, \{ (\neg H, \neg H)\})$ & \ref{it:r2.1} \\
    11 & 0	 & $\neg H$ &	$\TT{good boy!}$ &			$S$ &		$(\TT{good boy!}, \{ (\neg H,  S)^* \})$ & \ref{it:r1.2} \\
    12 & 1	 & $S$ &	\TT{I am going to grandma} &	$\neg H$ &		$(\TT{I am going to grandma}, \{ (S, \neg H)\} )$ & \ref{it:r2.1} \\
    & 2 &  &	 &	 &		$(\TT{good boy!}, \{ (\neg H, \neg H) \})$ & \ref{it:r3} \\
    13 & 3	 & $\neg H$ &	\TT{I am going to grandma} &	$\neg H$ &		$(\TT{I am going to grandma}, \{ (\neg H, \neg H)\} )$ & \ref{it:r2.1} \\
  \hline
\end{tabular}
\end{table}

At iteration 10 the action state $\neg H$ accepts the empty utterance $\epsilon$ through application (rule \ref{it:r2.1}). Then the user utters the unknown \TT{good boy!}, thereby inducing a state transition $S = \Phi(\neg H)$ by virtue of the intrinsic periodic dynamics, which becomes learned according to rule \ref{it:r1.2}. However, for the user intends \TT{good boy!} to be pragmatically affirmative, she repeats \TT{I am going to grandma} at iteration 12 yielding $\neg H$ when applied to the present state $S$. Furthermore, \TT{good boy!} is revised to the intended corroborative meaning $(\neg H, \neg H)$ subsequently. Finally, at iteration 13, the user wants to ensure the last transition by another utterance of \TT{I am going to grandma} which is accepted by the epistemic state $\neg H$, thus preventing any further revisions.

Eventually the lexicon
\begin{multline*}
	M_{14} = \{ (\epsilon, \{ (H, H), (\neg H, \neg H) \} ), \\
    (\TT{I am going to grandma}, \{ (H, \neg H), (\neg H, \neg H) , (S, \neg H) \} ) , \\
    (\TT{no!}, \{ (H, \neg H), (S, \neg H) \}), (\TT{heat!}, \{ (\neg H, \neg H)^* \}), \\
        (\TT{good boy!}, \{ (\neg H, \neg H) \} ) \} \:,
\end{multline*}
has been learned by virtue of Machine Semiotics.

\section{Discussion}
\label{sec:disc}

In this study we have proposed \emph{Machine Semiotics} as a viable approach to train speech assistive cognitive user interfaces in a perception-action cycle (PAC) between a user and a cognitive dynamic system \citep{Haykin12, Young10} through reinforcement learning of antecedent-consequent relations \citep{Skinner57, SuttonBarto18}. Our approach essentially rests on constructivist semiotics  \citep{MaturanaVarela98, Forster03} and biosemiotics \citep{Uexkull82}. Assuming the existence of species- and specimem-relative PACs for an individual in its ecological niche, its umwelt splits into a merkwelt, corresponding to its specific perception, and a wirkwelt, corresponding to its particular action faculties, in such a way that the individual's sensors provide only relevant information about the state of the environment, where relevance is characterized by the individual's action capabilities.

\Citet{Uexkull82} has nicely illustrated this idea by means of a tick, sitting on a tree branch and awaiting a mammal to attack. The tick has a largely reduced sensory system, basically comprising temperature, butyric acid smell, and fury touch. Thus, the meaning of ``mammal'' in the tick's merkwelt is essentially ``musty fury warmth''. Yet, every perception is attributed to a corresponding action: An increase of the butyric acid concentration in the air causes release from the branch to land in the prey's coat. Afterwards, the tick crawls toward the skin by avoiding the prey's hairs. Finally, maximal temperature induces the tick to sting the mammal's skin and sucking its blood. Thus, each percept becomes meaningfully relevant only in light of its corresponding action from the tick's wirkwelt. Correspondingly, the meaning of ``mammal'' in the tick's wirkwelt becomes the action sequence: release,  crawl, and sting.

Applied to a toy model of a cognitive heating \citep{KlimczakWolffLindemannEA14, HuberEA18}, the merkwelt is constituted by temperature sensation in combination with a basic symbolic language representation \citep{GravesMohamedHinton13, SundermeyerNeySchluter15} (where we deliberately have neglected its peculiar intricacies here). On the other hand, the wirkwelt of the device comprises (in our first scenario) only two possible actions: to heat or not to heat, together with displaying the current operation mode to the user, thereby providing observability of the machine states \citep{RussellNorvig10}. Following the constructivist approach of \citet{MaturanaVarela98}, that linguistic behavior of a sender aims at changing the behavior of the receiver, we interpret the meaning of the users' linguistic behaviors as mappings from antecedent onto consequent action states. This so-called ABC schema of verbal behavior \citep{Skinner57} forms the foundation of dynamic semantics \citep{Gardenfors88, GroenendijkStokhof91, Kracht02, Graben14a} and interactive pragmatics \citep{Grice89, AtlasLevinson81, Horn84, Blutner98b, Blutner06, Benz16, RooijJager12, HawkinsFrankeEA22}.

Accordingly, our Machine Semiotics substantially utilizes the basic tenets of dynamic semantics. The meaning of an utterance is a partial function upon a space of epistemic action states. Fixed points of those meaning functions define the important concept of acceptance. Our training algorithm describes the acquisition of semantic and pragmatic world knowledge from an ecological point of view \citep{Uexkull82}. Starting from two crucial assumptions, that (1) silence of the user endorses a current operation mode, and (2) linguistic behavior indicates disagreement with a current operation mode, we have proposed a reinforcement mechanism which continually revises inappropriate meanings upon upcoming new input. We have illustrated this algorithm in two case scenarios. The first one describes the two-state cognitive heating mentioned above where revision simply replaces a consequent state by its complement while keeping the antecedent. In a slightly more involved three-state scenario, we have employed an intrinsic periodic update dynamics. In both settings, we have abstractly represented the meaning of a user's utterance in the heater's wirkwelt through an action of its epistemic state space. Moreover, we have demonstrated that the algorithm is likely to converge when the user complies with the normative Cooperative Principle and the Conversation Maxims, originally suggested by \citet{Grice89} and later formalized in the framework of formal pragmatics and optimality theory \citep{AtlasLevinson81, Horn84, Blutner98b, Blutner06, Benz16, HawkinsFrankeEA22}.

Although Machine Semiotics is explicitly devised as a machine learning attempt for artificial intelligence, one might speculate about its tentative relevance for human language acquisition and semiosis in general. According to \citet{Harnad90} the ``symbol grounding problem'' for assigning symbolic ``meaning'' to almost continuous perceptions could be solved by means of ``discrimination and identification''. The former process creates ``iconic representations'' by training classifiers; the latter gives rise to conceptual categorizations. Yet these processes emphasize the perceptional aspects of the perception-action cycle. On the other hand, \citet{Steels06} argued that also the action space of embodied agents must be taken into account. This is actually achieved by Machine Semiotics.

Finally, we state that the Machine Semiotics outlined sofar is only able to acquire machine-relevant meanings only in form of instructions that aim at directly changing the behavior of the receiver \citep{Grice89, MaturanaVarela98, Forster03}. In terms of \citet{Dennett89g} and \citet{Posner93} these refer to \emph{first-order intentional systems} or the \emph{first level of reflection}, respectively. Here, ``a \emph{first-order} intentional system has beliefs and desires (etc.) but no beliefs and desires \emph{about} beliefs and desires'' \citep[p. 243]{Dennett89g}. Correspondingly, a system at the first level of reflection either ``beliefs something [\dots] or intends something.'' \citep[p. 227]{Posner93}. Yet consider a user who wants the machine to believe some proposition that is not directly related to its action. This would require a second-order intentional system which ``has believes and desires [\dots] about beliefs and desires'' \citep[op. cit.]{Dennett89g}, or equivalently, on the second level of reflection, ``$a$ believes that $b$ intends $a$ to believe something, and $b$ intends $a$ to believe that $b$ intends something.'' \citep[op. cit.]{Posner93}.  A Machine Semiotics that is able to deal with such scenarios would definitely demand a recursively structured epistemic state space together with a \emph{Theory of Mind} about the user's epistemic states \citep{Grice89}.

\section*{Acknowledgements}

We thank Matthias Wolff and Martin Wheatman for inspiring discussions.



\end{document}